\documentclass[runningheads]{llncs}
\usepackage{pdflscape} % or \usepackage{lscape} for DVI output
\usepackage{hyperref}
\usepackage{comment}
\usepackage[
backend=biber,
style=alphabetic,
sorting=ynt
]{biblatex}
\usepackage{amsmath}
\usepackage{booktabs} % For formal tables
\usepackage{graphicx}
\begin{document}
\title{General surgery vision transformer: A video pre-trained foundation model for general surgery}
\titlerunning{General surgery vision transformer}
%\author{Anonymous Authors}
\author{Samuel Schmidgall\inst{1}\orcidID{0000-0001-8192-9337} \and
Ji Woong Kim\inst{1}\orcidID{0000-0001-8669-205X} \and
Jeffrey Jopling\inst{1}\orcidID{0000-0002-0694-1704} \and
Axel Krieger\inst{1}\orcidID{0000-0001-8169-075X}}
\authorrunning{Schmidgall et al.}
\institute{Johns Hopkins University, Baltimore MD 21218, USA
\\ \email{sschmi46@jhu.edu}}
\maketitle              % typeset the header of the contribution
\begin{abstract}
The absence of openly accessible data and specialized foundation models is a major barrier for computational research in surgery. Toward this, (i) we open-source the largest dataset of general surgery videos to-date, consisting of 680 hours of surgical videos, including data from robotic and laparoscopic techniques across 28 procedures; (ii) we propose a technique for video pre-training a general surgery vision transformer (GSViT) on surgical videos based on forward video prediction that can run in real-time for surgical applications, toward which we open-source the code and weights of GSViT; (iii) we also release code and weights for procedure-specific fine-tuned versions of GSViT across 10 procedures; (iv) we demonstrate the performance of GSViT on the Cholec80 phase annotation task, displaying improved performance over state-of-the-art single frame predictors.
The code and data is available at: \href{https://github.com/SamuelSchmidgall/GSViT}{https://github.com/SamuelSchmidgall/GSViT}.

\end{abstract}
\section{Introduction}

Great progress has been made in Artificial Intelligence (AI) through the development of reusable general-purpose models.
These models are characterized by their large-scale training on extensive and diverse datasets, which enables them to develop a broad understanding of various subjects and skills. 
The concept of a \textit{foundation model} suggests a general base model that can be adapted or fine-tuned for specific applications, making them highly versatile for downstream tasks.
Publicly releasing foundation models and their respective datasets has largely been responsible for the incredible progress of language and vision models in recent years.

While progress toward building foundation models for medical applications has been steady, it has also advanced less quickly than applications outside of medicine.
A factor contributing to the slower progress, particularly in areas like surgery, is due to the immense volume of data required for training these models. 
Foundation models typically necessitate training on an enormous scale, ranging from tens of millions to trillions of samples. 
This scale poses significant challenges in medical applications, where data is subject to strict legal and ethical considerations.
The process of aggregating such a vast amount of patient data for training is challenging, and the process of making such data \textit{publicly accessible} is even more challenging.

We believe a step forward for the field of surgical AI is toward building a foundation model for general surgery.
Toward this, we introduce the General Surgery Vision Transformer (\texttt{GSViT}).
Several principles guided the design of our model architecture: (i) we aim to establish a parametrically lightweight foundation model such that it can be used in \textit{real-time} for real surgical applications, e.g. real-time surgeon feedback or robotic control; (ii) we aim to establish a model which has high compatibility with widely used models, noting that many existing vision-based transformer architectures utilize pre-trained convolutional networks as a pre-processing step; (iii) we aim to establish a model which is pre-trained from \textit{video prediction}, which we believe develops useful priors for the nature of surgical operations, encoding both spatial and temporal information which is useful for understanding tissue deformation.

We also introduce the GenSurgery dataset, a collection of surgical procedure videos which was used to pre-train \texttt{GSViT}.
This is the largest dataset in the field of surgery with 680 hours of general surgery demonstrations from robotic and laparoscopic techniques.
We avoid issues with confidentiality by curating our dataset from public sources, avoiding concerns of data sensitivity.
We release this dataset publicly to the surgical AI research community, aiming to accelerate research in the field with increased access to data.

%Several works have noted the inefficiency of non-medical models for medical tasks in 

\begin{figure}
    \centering    \includegraphics[width=0.99\textwidth]{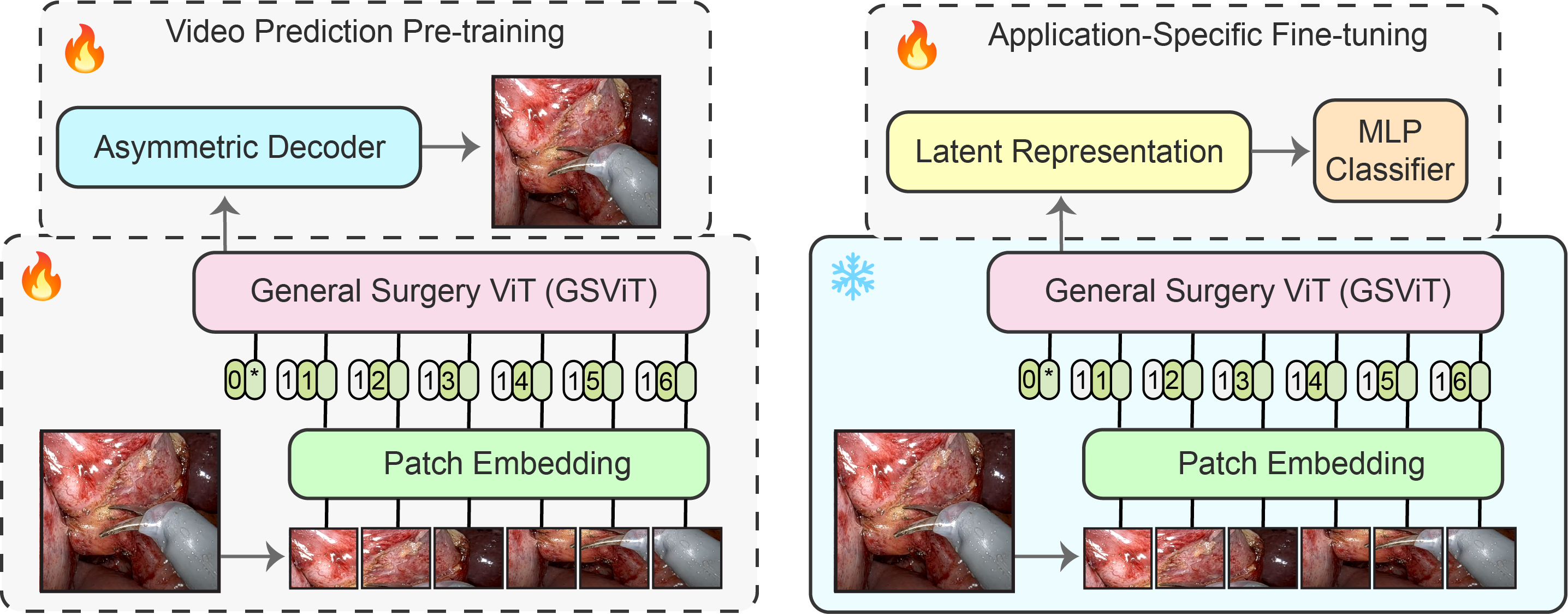}
    \caption{Graphical depiction of the training process for \texttt{GSViT}. Video prediction-based pre-training with asymmetric decoder head for video frame reconstruction (right). Demonstration of application-specific fine-tuning using \texttt{GSViT} with frozen weights and learned classification head (left).}
    \label{fig:final_results}
\end{figure}

\section{Related work}

\subsection{Foundation models in medicine}

In medicine, foundations models have been developed for applications such as radiology, endoscopic video analysis, and x-ray diagnoses.
These foundation models can largely be categorized into three architectures: language models, vision models, and vision-language models.
Language models are used in settings where diagnoses occur through dialogue, such as case studies presented from medical exams \cite{singhal2023towards,schmidgall2024addressing,ziaei2023language,nori2023can} or extracting data from clinical notes \cite{agrawal2022large}. 
Vision models are useful for medical tasks requiring visual perception, such as medical segmentation \cite{chen2021transunet,ma2024segment} (e.g. organs, tumors, medical tools) or disease classification \cite{zhou2023foundation}.
Vision-language models combine both vision and language, enabling the interaction of visual input with natural language, such as asking questions about images of a disease \cite{huang2023visual,li2023llava}.
Vision-based foundation models could be particularly impactful, as medical images are high dimensional and more challenging to acquire at volume, with the foundation model enabling medical tasks to be solved with less data.

\subsection{Vision transformers}

Vision Transformers (ViTs) extend the transformer architecture--initially designed for natural language processing tasks--to computer vision tasks.
The principle behind ViTs is to treat images as sequences of patches, akin to words in a sentence, thereby enabling the application of self-attention mechanisms to capture long-range dependencies within an image.

The architecture of a ViT begins with the division of an input image into a grid of fixed-size patches (see Fig. 1).
These patches are then linearly embedded into vectors of a specified dimension. 
An additional learnable embedding, often referred to as the "class token", is appended to these patch embeddings to represent the entire image. 
Positional embeddings are also often added to retain spatial information, as the transformer architecture does not inherently process sequential data in order.

More formally, given an input image \(X\), it is divided into \(N\) patches \\ \(\{P_1, P_2, \ldots, P_N\}\), where each patch \(P_i\) is transformed into a vector \(v_i\) through a linear projection:
\[ v_i = P_iW + b \]
where \(W\) and \(b\) are the weight matrix and bias vector of the linear projection layer, respectively. 
These vectors are then concatenated with a class token \(C\) and added to positional embeddings \(E_{pos}\) to form the input sequence for the transformer:
\[ Z_0 = [C; v_1; v_2; \ldots; v_N] + E_{pos} \]

The core of the ViT architecture is a series of transformer blocks, each comprising two main components: a multi-head self-attention (MSA) layer and a position-wise MLP. 
Each block processes the sequence as follows:
\[ Z'_l = \text{MSA}(\text{LN}(Z_{l-1})) + Z_{l-1} \]
\[ Z_l = \text{MLP}(\text{LN}(Z'_l)) + Z'_l \]
where \(Z_l\) is the output of the \(l\)-th transformer block, \(LN\) denotes layer normalization, and \(l\) ranges from 1 to the total number of blocks \(L\). 
The self-attention mechanism within MSA allows each patch to interact with every other patch, capturing global dependencies across the entire image.
After the final transformer block, the class token \(C\) is extracted and passed through a classification head, typically a linear layer, to produce the final output.

\subsection{Surgical Datasets}

Tool classification has some of the largest datasets in surgery \cite{twinanda2016endonet, al2019cataracts}, ranging from 86k to 900k frames.
There also exists a variety of surgical datasets focused on tool segmentation from various operations \cite{sznitman2012data, allan20202018, hong2020cholecseg8k}, typically ranging in size from 100-10K images.
Perhaps the largest surgical dataset is SurgToolLoc with 44M frames \cite{zia2023surgical}, focusing on tool presence classification.
Our work focuses on unsupervised pre-training, which most closely related to work done in endoscopy \cite{batic2023whether} and with MRI data \cite{prabhakar2024vit}.
We note that these models were trained on relatively small datasets compared to our work (307 and 700K images compared to 70M images).

\begin{figure}
    \centering
    \includegraphics[width=0.99\textwidth]{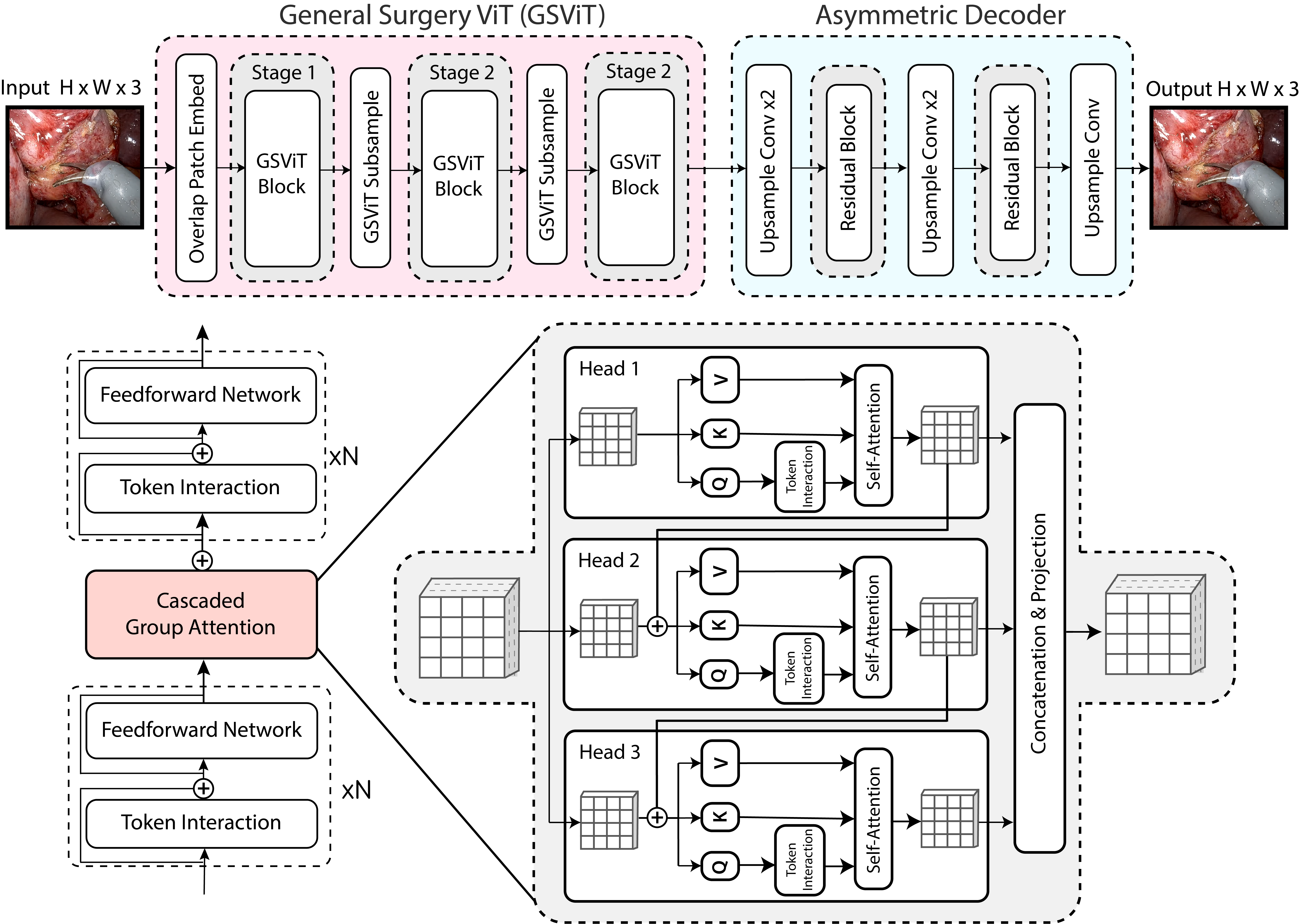}
    \caption{Architecture diagram of \texttt{GSViT}.}
    \label{fig:final_results}
\end{figure}

\section{Methods}

%\subsection{What are meaningful representation in medical applications?}

%Often in visual learning tasks, images are reduced to a lower dimensional latent space through the use of pre-trained vision models that have "frozen" weights (without additional weight tuning), such as Resnet-18 or ViT.
%The features extracted by these models are optimized primarily for image classification on the ImageNet dataset, which contains images of common objects such as cars and animals, resulting in representations that are optimized for tasks \textit{resembling ImageNet}. 
%However, this specialization can limit their effectiveness in visual tasks that require different kinds of image understanding \cite{sharma2022lossless}, such as object detection, image segmentation, or tasks in domains significantly different from those represented in the ImageNet dataset.
%This deviation becomes particularly noticeable in medical applications \cite{manzari2023medvit, ma2024segment}, where the salient feature space can deviate significant from common objects. 

\subsection{GSViT: General Surgery Vision Transformer}

The architecture we chose for \texttt{GSViT} is based on the EfficientNet ViT structure \cite{liu2023efficientvit}. 
The design of this model emphasizes speed in inference, minimizing memory and computation during inference and training time. 
We believe this is important for surgical applications, where algorithms run in real-time.
Below, we describe this architecture in detail (also see Fig. 1):

The sandwich layout is a unique structure where a single self-attention layer \( \Phi_A^i \) is placed between multiple Multi-Layer Perceptron (MLP) layers \( \Phi_F^i \). This configuration is formulated as:
\[ X_{i+1} = \Phi_F^N(\Phi_A^i(\Phi_F^N(X_i))) \]
where \( X_i \) is the input feature for the i-th block, and \( N \) denotes the number of MLP layers. This arrangement, with more MLP layers and fewer self-attention layers, aims to reduce memory consumption and facilitate channel communication.

In the Cascaded Group Attention (CGA) module, attention head redundancy is addressed by feeding each head with different splits of the full feature set, allowing for more efficient computation. This is given by the following formulation:
\[ X_{e_{ij}} = \text{Attn}(X_{ij}W_{Q_{ij}}, X_{ij}W_{K_{ij}}, X_{ij}W_{V_{ij}}) \]
\[ X_{e_{i+1}} = \text{Concat}[X_{e_{ij}}]_{j=1}^h W_{P_i} \]
where \( X_{ij} \) is the j-th split of the input feature \( X_i \), and \( W_{Q_{ij}}, W_{K_{ij}}, W_{V_{ij}} \) are the projection layers for each head. The cascaded design involves progressively refining the feature representation by adding the output of each head to the next. This is mathematically represented as:
\[ X'_{0_{ij}} = X_{ij} + X_{e_{i(j-1)}} \]
for \( 1 < j \leq h \). This approach saves computational resources and increases the depth of the feature representations.

Parameter reallocation within the network also plays an important role in memory conservation. This step involves redistributing parameters by expanding channel width in critical modules while reducing it in less important ones. This strategy allows for more efficient learning of representations in higher-dimensional spaces without losing feature information, while also speeding up inference by removing redundant parameters.

\subsubsection{Asymmetric Decoder}

The architecture is structured to progressively reconstruct high-resolution images from the \texttt{GSViT} encoded representation. 
It begins with an initial representation stage where a fully connected layer transforms the input into a higher-dimensional space, followed by batch normalization and activation using the Gaussian Error Linear Unit. 
This process prepares the encoded data for the decoding sequence by expanding its dimensions and stabilizing the learning process through normalization.

In the decoder, a sequence of transposed convolutional layers incrementally increases the spatial dimensions of the processed data to reconstruct the original image size. 
Each transposed convolutional layer is followed by batch normalization and a ReLU activation. 
The architecture employs varying sizes of filters and strides to control the upscaling process. 
In addition to basic decoding layers, the architecture integrates residual connections with Squeeze-and-Excitation (SE) attention modules at two different scales, aimed at refining feature representations by recalibrating channel-wise feature responses, thereby enhancing the model's ability to focus on relevant features during reconstruction.

\subsubsection{Pre-training through video prediction}

Unlike the applications of other medical foundation models which were trained for image reconstruction, surgery is spatial and \textit{temporal}. 
The majority of image-pretraining techniques, such as direct image reconstruction or masking techniques, train on a single image at an instance in time. 
However, we believe that the latent space embedding for surgical problems should capture both spatial and temporal properties and require developing models that account for both space and time.  
Recent work has explored pre-training using video prediction \cite{ gupta2022maskvit}, showing improved results for spatio-temporal problems. 
We take a similar approach \texttt{GSViT}, performing next-frame reconstruction for pre-training with a 1 frame input and a 1 second prediction (1 frame).
We found that the addition of masking, such as in \cite{gupta2022maskvit}, did not improve performance and opted for the next-frame reconstruct as demonstrated in \cite{gao2022simvp}.

\section{Results}

\subsection{Creating GenSurgery from public YouTube videos}

We used publicly available videos of laparoscopic and robotic surgery to establish the largest surgical dataset for computer vision.
For the data collection process, a selection of videos was curated from specific playlists of surgical demonstrations on YouTube using surgery related keywords. %(see \textit{list of keywords}). 
The video channels were recorded from each of these playlists, and, following this, additional videos were obtained from the creators of the original video sets.
After data collection, duplicates were programmatically removed and every video was manually screened to ensure the content was a surgical demonstration rather than other common content on surgery, e.g. lecture videos.

Further, each video was categorized into the specific operation that was performed as well as whether it was a laparoscopic or robotic surgery.
The number of videos for each type of operation is demonstrated visually in Figure 2, as well the number of frames total for each type of operation.
We note that the operation type with the largest number of frames versus total videos differs quite significantly because operations can vary significantly in time (e.g. hernia repair averages 90.8 minutes \cite{qabbani2021robotic} and hepatectomy averages 424.4 minutes \cite{jang2022early}). 
In total, GenSurgery has a total of 70M frames.

%Data was recorded at the highest available frame rate and at the highest available resolution to allow for a wider range of applications of the data.

\begin{figure*}[ht]
    \centering    \includegraphics[width=0.99\textwidth]{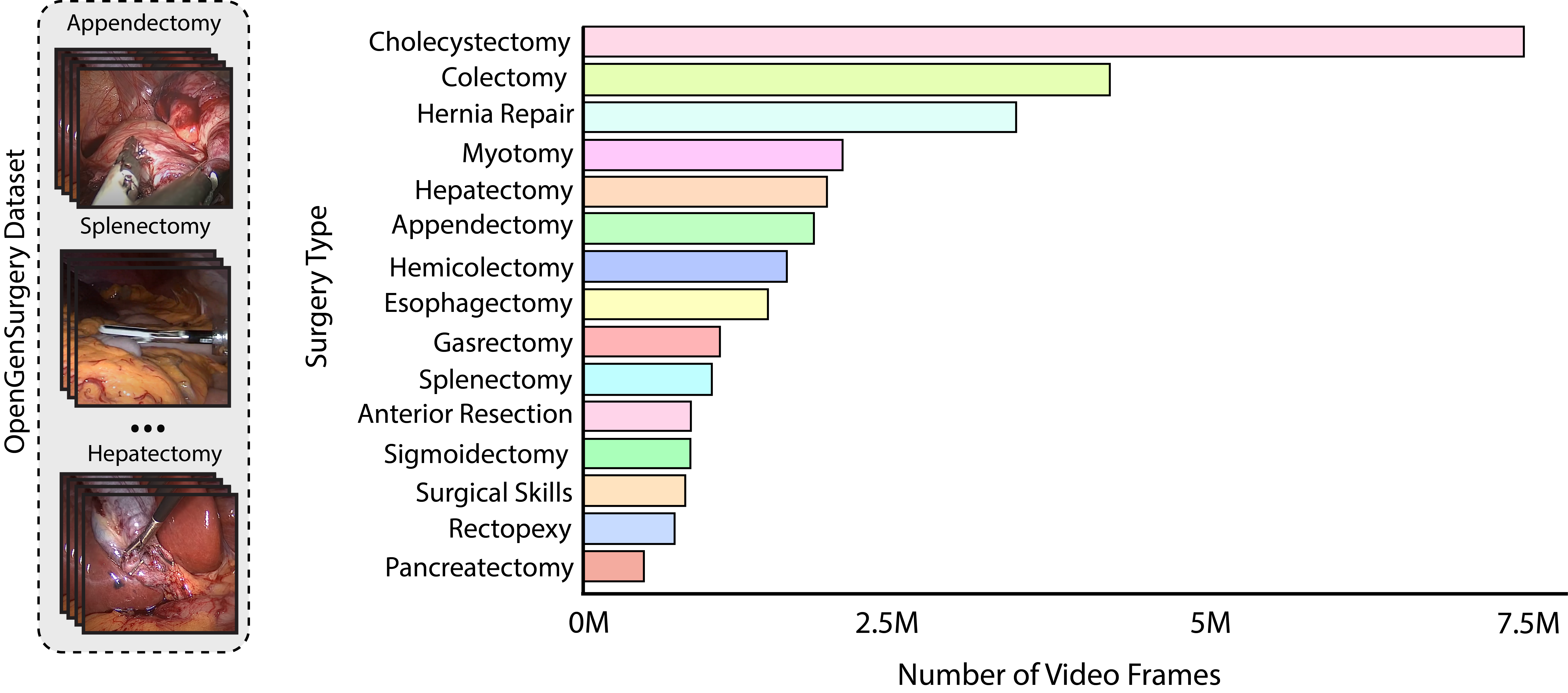}
    \caption{GenSurgery.}
    \label{fig:final_results}
\end{figure*}

\subsection{Fine-tuning on surgical procedures}

In addition to the base foundation model \texttt{GSViT}, we also release fine-tuned models for ten types of surgical procedures.
These were chosen based on having the largest number of frames in GenSurgery.
Fine-tuning followed the same procedure as pre-training except that next-frame prediction was only optimized on data from particular procedures.
We produce models for the following operations: Appendectomy, Cholecystectomy, Colectomy, Gastrectomy, Hernia Repair, Esophagectomy, Colostomy, Hemicolectomy, Hepatectomy, Splenectomy.
Model are named with the following structure: \texttt{GSViT-SurgeryType} (e.g. \texttt{GSViT}-\texttt{Colostomy}.
Each model was fine-tuned for one epoch across the dataset with a learning rate of \texttt{1e-4}.

\subsection{GSViT can run in real-time applications}

Unlike other areas of medicine, surgery-aided tools must be capable of running in real-time as the surgeon is operating.
As mentioned, the design of GSViT took this into consideration, and was optimized for compute efficiency.
Briefly, we demonstrate results for run-time performance metrics of \texttt{GSViT} compared with alternative architectures.

We compare the runtime inference time (no gradient accumulation) for each the of following models on input images from the Cholec80 dataset.
Here, we note that \texttt{GSViT} is able to process 10621 images per second \cite{liu2023efficientvit} which is 10.6 images per millisecond. 
In a non-parallelized process we attain a runtime average of $12.1\pm0.1$ ms on a single 12GB NVIDIA RTX A5500 GPU.
We can compare this with two state-of-the-art efficient architectures, EfficientNet-B0 \cite{tan2019efficientnet} and GLiT-Tiny \cite{chen2021glit}, which attain a throughput average of $4532$ and $3516$ images per second (inference time of $4.5$ and $3.5$ images per millisecond) respectively.

\subsection{Surgical phase detection}

We compare the performance of \texttt{GSViT} on the Cholec80 surgical phase detection benchmark.
We adapt the fine-tuned \texttt{GSViT-Cholecystectomy} for classification, starting with detaching the asymmetric decoder from the pre-training phase and append an MLP classifier to the model, taking the \texttt{GSViT} latent representation as input (see Fig. 1). 
Attached are three MLP layers with ELU activations followed by batch normalization.
During training, we use dropout to improve generalization at each of the hidden layers with a $10\%$ probability.
We also introduce several image transformations during training to augment the dataset, such as photometric distortion and Gaussian blurring (see code repository for a comprehensive list).

We use a learning rate of $3e-4$ with exponential decay and a batch size of $128$.
The MLP classifier head outputs a $7$ dimensional classification vector and has two layers with $2048$ and $512$ hidden neurons respectively. 
The hidden neurons have exponential linear unit (ELU) activations \cite{clevert2015fast} followed by a layer normalization \cite{ba2016layer}.

%\begin{table}[ht]
%\centering
%\caption{Model performance on Cholec80 Surgical Tool Annotation task.}
%\label{your-table-label}
%\begin{tabular}{lcccc}
%\toprule
%\textbf{Model name } & \textbf{\hspace{9pt} Accuracy \hspace{9pt}} & \textbf{\hspace{6pt} Precision \hspace{6pt}} & \textbf{\hspace{6pt} Recall \hspace{6pt}} & \textbf{ Params } \\
%\midrule
%PhaseNet \cite{twinanda2016endonet} & $78.8 \pm 4.7$ & $71.3 \pm 15.6$ & $76.6 \pm 16.6$ & 58.3M \\ 
%EndoNet \cite{twinanda2016endonet} & $81.7 \pm 4.2$ & $73.7 \pm 16.1$  & $79.6 \pm 7.9$ & 58.3M \\
%OHFM \cite{yi2019hard} & $87.3 \pm 5.7$ & --- & --- & 47.1M \\
%TeCNO \cite{czempiel2020tecno} & $88.6 \pm 7.8$ & $86.5 \pm 7.0$ & $87.6 \pm 6.7$ & 24.7M \\
%MTRCNet-CL \cite{jin2020multi} & $89.2 \pm 7.6$ & $86.9 \pm 4.3$ & $88.0 \pm 6.9 $ & 29.0M \\ 
%Trans-SVNet & $90.3 \pm 7.1$ & $90.7 \pm 5.0$ & $88.8 \pm 7.4$ & 24.7M \\
%\midrule
%\texttt{GSViT-Cholecystectomy} & $86.3 \pm 5.9$ & $82.5 \pm 7.1$ & $84.7 \pm 5.3$ & \textbf{13.7M} \\ %26.1M
%\bottomrule
%\end{tabular}
%\end{table}

\begin{table}[ht]
\centering
\caption{Model performance on Cholec80 Surgical Phase Annotation task.}
\label{your-table-label}
\begin{tabular}{l|cccccc}
\toprule
\textbf{Metric \hspace{18pt}} & \textbf{ \hspace{3pt}PhaseNet \hspace{3pt}} & \textbf{ \hspace{3pt}EndoNet \hspace{3pt}} & \textbf{ \hspace{3pt}TeCNO \hspace{3pt}} & \textbf{ \hspace{3pt}LoViT \hspace{3pt}} & \textbf{ \hspace{2pt}GSViT (ours)\hspace{2pt}} \\
\midrule
\textbf{Accuracy } & $78.8 \pm 4.7$ & $81.7 \pm 4.2$ & $88.6 \pm 7.8$ & $92.4 \pm 6.3$ & $86.3 \pm 5.9$ \\
\textbf{Precision } & $71.3 \pm 15.6$ & $73.7 \pm 16.1$ & $86.5 \pm 7.0$ & $ 89.9 \pm 6.1$ & $82.5 \pm 7.1$ \\
\textbf{Recall } & $76.6 \pm 16.6$ & $79.6 \pm 7.9$ & $87.6 \pm 6.7$ & $90.6 \pm 4.4$ & $84.7 \pm 5.3$ \\
\textbf{Params} & 58.3M & 58.3M & 24.7M & 29.0M & 13.7M
\\
\textbf{Frames} & 1 & 1 & All & 100, 500 & 1
\\
\bottomrule
\end{tabular}
\end{table}

\begin{comment}
\begin{table}[ht]
\centering
\caption{Model performance on Cholec80 Surgical Tool Annotation task.}
\label{your-table-label}
\begin{tabular}{lcccc}
\toprule
\textbf{Model name } & \textbf{\hspace{9pt} Accuracy \hspace{9pt}} & \textbf{\hspace{6pt} Precision \hspace{6pt}} & \textbf{\hspace{6pt} Recall \hspace{6pt}} & \textbf{ Params } \\
\midrule
PhaseNet \cite{twinanda2016endonet} & $78.8 \pm 4.7$ & $71.3 \pm 15.6$ & $76.6 \pm 16.6$ & 58.3M \\ 
EndoNet \cite{twinanda2016endonet} & $81.7 \pm 4.2$ & $73.7 \pm 16.1$  & $79.6 \pm 7.9$ & 58.3M \\
OHFM \cite{yi2019hard} & $87.3 \pm 5.7$ & --- & --- & 47.1M \\
TeCNO \cite{czempiel2020tecno} & $88.6 \pm 7.8$ & $86.5 \pm 7.0$ & $87.6 \pm 6.7$ & 24.7M \\
MTRCNet-CL \cite{jin2020multi} & $89.2 \pm 7.6$ & $86.9 \pm 4.3$ & $88.0 \pm 6.9 $ & 29.0M \\ 
%Trans-SVNet & $90.3 \pm 7.1$ & $90.7 \pm 5.0$ & $88.8 \pm 7.4$ & 24.7M \\
\midrule
\texttt{GSViT} (Ours) & $86.3 \pm 5.9$ & $82.5 \pm 7.1$ & $84.7 \pm 5.3$ & \textbf{13.7M} \\ %26.1M
\bottomrule
\end{tabular}
\end{table}
\end{comment}

The results of \texttt{GSViT-Cholecystectomy} trained on Cholec80 is shown in \textbf{Table 1}.
We demonstrate comparable performance to previous models, obtaining $86.3\%$ accuracy, at a significantly reduced computational cost and a reduced amount of tuneable parameters\footnote{The total number of parameters is 26.1M, with 13.7M tuneable parameters and an additional 12.4M non-tuneable parameters for the foundation model}. 
We also note that our approach recognizes phase by only looking at the current frame, whereas other approaches such as TeCNO \cite{czempiel2020tecno} take in the entire video history for a phase classification.
Considering this, our model is the highest performance single-frame model.
We incorporate this to make it more feasible for real-time surgical applications.

\section{Discussion}

In this work, we introduce the General Surgery Vision Transformer (\texttt{GSViT}), which is a parametrically efficient pre-trained vision transformer, trained for video prediction on millions of surgical video frames. 
We also introduce the GenSurgery dataset, comprising 680 hours of surgery videos from 28 different procedures.
We open-source the code and weights for \texttt{GSViT} as well as fine-tuned versions for 10 different procedures.
We show that \texttt{GSViT} is able to run in real-time while getting comparable performance on the Cholec80 surgical phase classification task.

Future work could expand the pre-training by incorporating temporal and spatial mask reconstruction.
Additionally, methods for obtaining higher performance on the Cholec80 dataset could be explored such as temporal convolutions \cite{czempiel2020tecno}, long videos \cite{liu2023lovit} or recurrent convolutions \cite{jin2020multi}.
We also note that our dataset could be used for building a more general-purpose medical foundation model that includes surgery as well as additional modalities such as MRI and CT images \cite{schmidgall2024general, schmidgall2024robots}.

We believe that surgical AI has the potential to shape the future of surgery and hope that this work is a step towards that grand vision.

\section{Acknowledgement}

This material is based upon work supported by the National Science Foundation Graduate Research Fellowship for Comp/IS/Eng-Robotics under Grant No. DGE 2139757 and NSF/FRR 2144348 and ARPA-H 75N91023C00048.

\printbibliography

\end{document}